\documentclass{article}
\usepackage{spconf,amsmath,graphicx,color}

\title{RADNET: Radiologist Level Accuracy using Deep Learning for HEMORRHAGE detection in CT Scans}
\name{Monika Grewal, Muktabh Mayank Srivastava, Pulkit Kumar\textsuperscript{*}, Srikrishna Varadarajan\sthanks{Authors contributed equally}}
\address{ParallelDots, Inc.\sthanks{www.paralleldots.xyz}\\
{\tt\small{\{monika, muktabh, pulkit, srikrishna\}@paralleldots.com }}}

\begin{document}
%
\maketitle
\begin{abstract}
We describe a deep learning approach for automated brain hemorrhage detection from computed tomography (CT) scans. Our model emulates the procedure followed by radiologists to analyse a 3D CT scan in real-world. Similar to radiologists, the model sifts through 2D cross-sectional slices while paying close attention to potential hemorrhagic regions. Further, the model utilizes 3D context from neighboring slices to improve predictions at each slice and subsequently, aggregates the slice-level predictions to provide diagnosis at CT level. We refer to our proposed approach as Recurrent Attention DenseNet (RADnet) as it employs original DenseNet architecture along with adding the components of attention for slice level predictions and recurrent neural network layer for incorporating 3D context. The real-world performance of RADnet has been benchmarked against independent analysis performed by three senior radiologists for 77 brain CTs. RADnet demonstrates 81.82\% hemorrhage prediction accuracy at CT level that is comparable to radiologists. Further, RADnet achieves higher recall than two of the three radiologists, which is remarkable.
\end{abstract}
\section{Introduction}
\label{sec:intro}

Computed tomography (CT) is the most commonly used medical imaging technique to assess the severity of brain hemorrhage in case of traumatic brain injury (TBI). The diagnosis of hemorrhage following TBI is extremely time-critical as even a delay of few minutes can cause loss of life. Traditional methods involve visual inspection by radiologists and quantitative estimation of the size of hematoma and midline shift manually. The entire procedure is time-consuming and requires the availability of trained radiologists at every moment. Therefore, automated hemorrhage detection tools capable of providing fast inference, which is also accurate to the level of radiologists; hold the potential to save thousands of patient lives.

Deep learning-based automated diagnosis approaches have been gaining interest in recent years, mainly due to their faster inference and ability to perform complex cognitive tasks, which otherwise required specialized expertise and experience. In this work, we have sought applicability of deep learning for hemorrhage detection from brain CT scans. Taking motivation from real-world where radiologists take decision for each slice by combining information from neighboring slices; we have modeled 3D CT labeling task as a combination of slice-level classification task and a sequence labeling task to incorporate 3D inter-slice context. This is done by combining Convolutional Neural Network (CNN) with Long-Short Term Memory (LSTM) network. We used DenseNet architecture \cite{Huang_2017_CVPR} as baseline CNN for learning slice-level features and bidirectional LSTM for combining spatial dependencies between slices. In addition, we used the prior information about hemorrhagic region of interest to focus DenseNet\textsc{'}s attention towards relevant features. We refer to our modified architecture as Recurrent Attention DenseNet (RADnet).

For any automated system to be deployable in clinical emergency set-up, reliable estimation along with high sensitivity to the level of human specialists is required. This necessitates the need to benchmark against specialists in the field. To this end, we have benchmarked the performance of RADnet against three senior radiologists.

In the following sections, we first go through the related work in the field. Section 2 describes the implementation details including architecture and training details of RADnet. Section 3 represents the comparison results between RADnet and three senior radiologists. The possible caveats in the proposed approach and required future work are discussed in section 4.
\begin{figure*}[t]
\centering
{\includegraphics[width=0.8\textwidth]{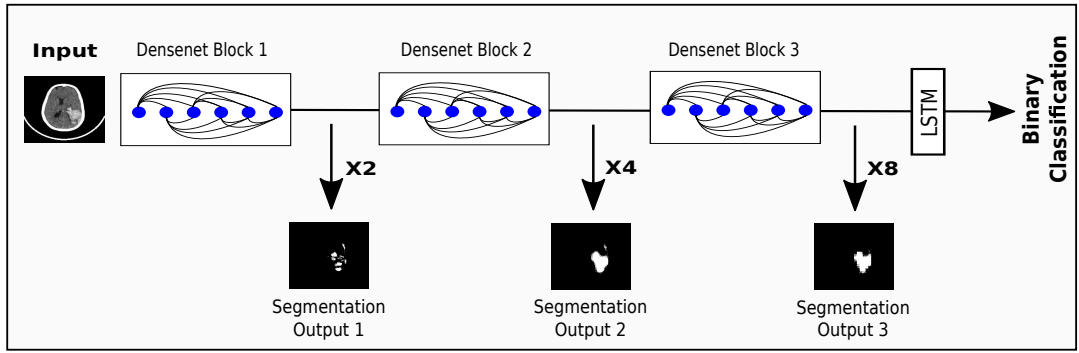}}
\caption{\label{Fig:figure-name}\footnotesize{Architecture details of RADnet. X2, X4, and X8 represent upsampling by a factor of 2, 4, and 8 using deconvolution layer}}
\centering
\end{figure*}

\subsection{Related work}
\label{ssec:subhead}

Many deep learning approaches have been proposed for medical image classification \cite{Gulshan2016DevelopmentPhotographs}\cite{Gao2017ClassificationNetworks}\cite{7401052}\cite{Wang2017Zoom-in-Net:Detectionc} as well as segmentation tasks \cite{Ronneberger2015U-Net:Segmentation}\cite{chen2017voxresnet}. Some studies have also reported quite promising results as compared to previously existing automated analysis approaches. However, only a few studies have benchmarked their methods against the accuracy of human specialists. Rajpurkar and Hannun et. al. have presented a 34-layer convolutional network that detects arrhythmia from electrocardiogram (ECG) data with cardiologists level accuracy \cite{Rajpurkar2017Cardiologist-LevelNetworks}. Another study has reported super-human accuracy in 3D neuron reconstruction from electron microscopic brain images \cite{Lee2017SuperhumanChallenge}. Lieman-Sifryet. al.  have recently reported a plausible real-world solution for automated cardiac segmentation \cite{Lieman-Sifry2017FastVentricle:ENetb}. Taking a step forward in the same direction, we have validated the performance of RADnet against annotations obtained from senior radiologists. To the best of our knowledge, this is the first work reporting a deep learning solution for brain hemorrhage detection that has been validated against human specialists in the field.

Several attention mechanisms have been previously utilized to increase focus of the neural network towards salient features \cite{Mnih2014RecurrentAttention}\cite{Chen2015ABC-CNN:Answering}. Existing attention mechanisms can be divided in the categories of hard and soft attention. Hard attention methods shift attention from one part of image to another part through sampling the images \cite{Ba2014MultipleAttention}. Whereas, soft attention methods utilize probabilistic attention maps to focus on certain features more than the others \cite{Zagoruyko2016PayingTransfer}, for instance, the features corresponding to specific task \cite{Chen2015ABC-CNN:Answering} or different scale objects \cite{Chen2015AttentionSegmentation}. The attention method described by us is slightly different from the existing approaches in the way that it makes use of prior information available in form of contours delineating hemorrhagic regions.

The combination of CNN architectures with LSTM has been extensively explored for tasks that require modeling long-term spatial dependencies within image e.g. image captioning \cite{Vinyals2014ShowGenerator} and action recognition \cite{Li2017Skeleton-basedNetworks}; or temporal dependencies between consecutive frames e.g. video recognition tasks \cite{donahue2015long}. Recently, a few studies on biomedical imaging have explored this architecture for leveraging inter-slice dependencies in 3D images \cite{2016arXiv160909143Y}\cite{chen2016combining}\cite{DBLPCaiLXXY17}. Our approach is more similar to the latter approaches in that it models 3D aspect of medical images as a sequence.

\section{MATERIALS AND METHODS}
\label{sec:format}
\subsection{Dataset}
\label{ssec:subhead}

The dataset composed of 185, 67, and 77 brain CT scans for training, validation, and testing respectively. The dataset were obtained from two local hospitals after the approval from ethics committee. The images were of varying in-plane resolutions (0.4 mm - 0.5 mm) and slice-thicknesses (1 mm - 2 mm). The training and validation CTs were annotated at slice-level for the class labels and segmentation contours delineating hemorrhagic region using in-house web-based annotation tool.  The testing data was independently annotated by three senior radiologists for the presence of hemorrhage at slice-level as well as at CT level. The reports of the patients, which were generated after consultation from two senior radiologists along with correlation to medical history were used as the ground truths for test data.

\subsection{Implementation details}
\label{ssec:subhead}
\subsubsection{Preprocessing}
\label{sssec:subsubhead}
The Hounsfield Units (HU) in raw CT scan images were thresholded to brain window and images were resampled to have isotropic (1 mm x 1 mm) resolution and 250 mm x 250 mm field-of-view in-plane. No resampling was performed along z-axis. The HU levels were then converted to the range 0 to 1.
\subsubsection{Architecture}
\label{sssec:subsubhead}
We have used 40-layer DenseNet \cite{Huang_2017_CVPR} architecture with three dense blocks as baseline network. Further we added three auxiliary tasks to compute binary segmentation of hemmorhagic regions. These auxiliary tasks allow the network to explicitly focus its attention towards hemorrhagic regions, which in turn, improves the performance on classification task. The auxiliary task branches were forked after the last concatenation layer in each dense block [Fig. 1] and consisted of single module of 1 x 1 convolutional layer followed by deconvolution layer to upsample the feature maps to original image size. The final loss was defined as the weighted sum of cross-entropy loss for main classification task and losses for three auxiliary tasks. We refer to this modification as DenseNet-A.

The final RADnet architecture models the inter-slice dependencies between 2D slices of CT scans by incorporating bidirectional LSTM layer to DenseNet-A. The output of the global pool layer in original architecture is passed through single bidirectional LSTM layer before sending to fully connected layer for final prediction. Instead of independent 2D slices, we trained the RADnet with sequences of multiple 2D slices. In this way, the network models the 3D context of CT scan in a small neighborhood, while predicting for single slice. The extent of 3D neighborhood was decided based on GPU memory constraints and to maximize performance on validation set. 

\subsubsection{Training}
\label{sssec:subsubhead}
The training was done using stochastic gradient descent for 60 epochs with initial learning rate of 0.001 and 0.9 momentum. The learning rate was reduced to 1/10\textsuperscript{th} of the original value after 1/3\textsuperscript{rd} and 2/3\textsuperscript{rd} of the training finished. The network weights were initialized using He norm initialization \cite{henorm}. The dataset was augmented using rotation and horizontal flipping to balance positive and negative class ratios along with increasing generalizability of the model. The hyperparameters were tuned so as to give best performance on validation set. Training took 40 hours on a single Nvidia GTX 1080Ti GPU.

\subsection{Evaluation}
\label{ssec:subhead}
The slice level predictions from RADnet were aggregated to infer CT level predictions. A 3D CT scan was declared positive if the deep learning model predicted hemorrhage in consecutive three or more slices. This corresponds to minimum cross-sectional dimension of approximately 1.5 mm for hemorrhagic region. Further, we calculated accuracy, recall, precision, and F1 score of the CT level predictions by RADnet and compared with the annotations from individual radiologists.
\begin{table}[!h]
\centering
\begin{tabular}{|l|c|c|c|c|}
\hline
& \textbf{DenseNet} & \textbf{DenseNet-A}\textsuperscript{a} & \textbf{RADnet}\\
\hline
\textbf{Accuracy} & 72.73\% & 80.52\% & \textbf{81.82}\%\\
\hline
\textbf{Recall} & 84.10\% & \textbf{90.91}\% & 88.64\%\\
\hline
\textbf{Precision} & 72.55\% & 78.43\% & \textbf{81.25}\%\\
\hline
\textbf{F1 score} & 77.90\% & 84.21\% & \textbf{84.78}\%\\
\hline
\end{tabular}
\centering
\caption{\footnotesize\label{tab:table-name}Performance evaluation of baseline DenseNet, DenseNet with attention, and RADnet architectures. \textsuperscript{a} Baseline Densenet with attention}
\end{table}

\section{RESULTS}
\label{sec:pagestyle}
Table 1 lists the evaluation metrics on test set for RADnet and two other baseline architectures: DenseNet and DenseNet with attention component. The comparison of metrics clearly highlights the advantage of adding the components of attention and recurrent network to original DenseNet architecture. Adding attention provided remarkable increase in both recall and precision. Whereas, adding recurrent LSTM layer provided a further increase in precision. The performance of final architecture, RADnet is compared with that of individual radiologists in Table 2. RADnet achieved 81.82\% accuracy, 88.64\% recall, 81.25\% precision, and 84.78\% F1 score. 
\begin{table}[!h]
\centering
\begin{tabular}{|l|c|c|c|c|}
\hline
 & \textbf{Rad. 1} & \textbf{Rad. 2} & \textbf{Rad. 3} & \textbf{RADnet}\\
\hline
\textbf{Accuracy} & 81.82\% & \textbf{85.71}\% & 83.10\% & 81.82\%\\
\hline
\textbf{Recall} & 81.82\% & \textbf{88.64}\% & 86.84\% & \textbf{88.64}\%\\
\hline
\textbf{Precision} & 85.71\% & \textbf{86.67}\% & 82.5\% & 81.25\%\\
\hline
\textbf{F1 score} & 83.72\% & \textbf{87.64}\% & 84.62\% & 84.78\%\\
\hline
\end{tabular}
\centering
\caption{\footnotesize\label{tab:table-name}Comparison between radiologists and RADnet. The dataset consisted of total 77 CT scans including 44 hemorrhage and 33 normal cases. \textit{Abbreviations}:- Rad. - Radiologist}
\end{table}
The accuracy of RADnet was same as one of the radiologists, and the recall of RADnet was higher than two radiologists. Further, RADnet showed higher F1 score than two of the three radiologists. The precision of RADnet predictions was slightly lower than the minimum precision of radiologists (RADnet precision - 81.25\%, minimum precision of radiologists - 82.5\%). However, it should be noted that the underlying task is more tolerant towards lower specificity as compared to lower sensitivity.

\section{DISCUSSION \& CONCLUSION}
\label{sec:typestyle}
We have presented a deep learning based approach RADnet that emulates radiologists' method for diagnosis of the brain hemorrhage from CT scans. The prediction results have been benchmarked against senior radiologists. RADnet achieves prediction accuracy comparable to the radiologists along with increased sensitivity. It is to be noted that high sensitivity is much-required characteristic of an automated approach to consider deployment as an emergency diagnostic tool. Further, we have presented an approach to use auxiliary task of pixel-wise segmentation as a method to focus deep learning model's attention towards relevant features for classification. This has an additional advantage that the obtained segmentation maps can be used as the qualitative indication of the severity of hemorrhage in the brain.

This is worth mentioning that although the proposed approach provides promising results with higher sensitivity than the radiologists in automated hemorrhage detection task, there still exist so many other equally severe brain conditions that the given deep learning model is unaware of. Future work in the direction of multiple pathologies detection from brain CT scans is therefore guaranteed. As such, the presented solution should not be misinterpreted as a plausible replacement for actual radiologists in the field. 

In summary, the proposed deep learning method demonstrates potential to be deployed as an emergency diagnosis tool. However, the method has been tested on a limited test set, and its real-world performance is still subjected to further experimentation.

\section{ACKNOWLEDGEMENTS}
\label{sec:print}
We thank the hospitals for providing data and the radiologists for providing annotations used as ground truth during training and comparison during testing. We would also want to thank the engineering team for developing the web-based annotation tool.
%
%

\bibliographystyle{IEEEbib}
\bibliography{ref}

\end{document}